\pdfoutput=1

\documentclass[11pt]{article}

\usepackage[]{EMNLP2022}
\usepackage{adjustbox}
\usepackage{times}
\usepackage{latexsym}
\usepackage{booktabs,subcaption,amsfonts,dcolumn}
\usepackage[T1]{fontenc}

\usepackage{hyperref}
\usepackage{inconsolata}

\usepackage{multirow}
\usepackage{times}
\usepackage{latexsym}
\usepackage[utf8]{inputenc}
\usepackage{xcolor,colortbl}
\usepackage{microtype}
\usepackage{url}
\usepackage{longtable}
\usepackage{tikz}
\usepackage{tabularx}
\usepackage{lscape}
\usepackage{hhline}
\usepackage{bbm}
\usepackage{array}
\usepackage{graphicx}
\usepackage{booktabs}
\usepackage{makecell}
\usepackage{amsfonts}       %
\usepackage{nicefrac}       %
\usepackage{microtype}      %
\usepackage{multirow}
\usepackage{caption}
\usepackage{amsmath}
\usepackage{float} 
\usepackage{ragged2e}
\usepackage{breqn}
\usepackage{soul}
\usepackage[T1]{fontenc}
\newcommand*\circled[1]{\tikz[baseline=(char.base)]{
            \node[shape=circle,draw,inner sep=2pt] (char) {#1};}}

\usepackage[utf8]{inputenc}

\usepackage{microtype}

\definecolor{LightCyan}{rgb}{0.91,0.96,0.97}
\definecolor{green}{rgb}{0.94,1,0.94}
\definecolor{grey}{rgb}{0.9,0.9,0.92}
\definecolor{Red}{rgb}{1,0.8,0.8}
\definecolor{viol}{rgb}{0.49, 0, 1}

\newcolumntype{b}{>{\columncolor{LightCyan}}c}
\newcolumntype{g}{>{\columncolor{grey}}c}
\newcolumntype{k}{>{\columncolor{green}}c}

\title{CORE: A Retrieve-then-Edit Framework \\ for Counterfactual Data Generation}

\author{
Tanay Dixit$^{1}$~ Bhargavi Paranjape$^{2}$ ~Hannaneh Hajishirzi$^{2,3}$ ~Luke Zettlemoyer$^{2,4}$ \\
$^{1}$ Indian Institute of Technology, Madras \\
$^{2}$ Paul G. Allen School of Computer Science \& Engineering, University of Washington  \\
$^{3}$ Allen Institute of Artificial Intelligence, Seattle ~~
$^{4}$ Meta AI \\
\texttt{tanay.dixit@smail.iitm.ac.in} \\
\texttt{\{bparan,hannaneh,lsz\}@cs.washington.edu}
}

\begin{document}
\maketitle
\begin{abstract}

Counterfactual data augmentation (CDA) -- i.e., adding minimally perturbed inputs during training -- helps reduce model reliance on spurious correlations and improves generalization to out-of-distribution (OOD) data. Prior work on generating counterfactuals only considered restricted classes of perturbations, limiting their effectiveness.
We present \textbf{CO}unterfactual Generation via \textbf{R}etrieval and \textbf{E}diting (\textbf{CORE}), a retrieval-augmented generation  framework for creating \emph{diverse} counterfactual perturbations for CDA. For each training example, CORE first performs a dense retrieval over a task-related unlabeled text corpus using a learned bi-encoder and extracts relevant counterfactual excerpts. CORE then incorporates these into prompts to a large language model with few-shot learning capabilities, for counterfactual editing. Conditioning language model edits on naturally occurring data results in diverse perturbations. Experiments on  natural language inference and sentiment analysis benchmarks show that CORE counterfactuals are more effective at improving generalization to OOD data compared to other DA approaches. We also show that the CORE retrieval framework can be used to encourage diversity in manually authored perturbations \footnote{Code at \url{https://github.com/tanay2001/CORE}}.

\end{abstract}


\section{Introduction}

Contrast sets \citep{gardner-etal-2020-evaluating} and counterfactual data \citep{kaushik2020learning} provide minimal input perturbations that change model predictions, and serve as an effective means to evaluate brittleness to  out-of-distribution data \citep{wang2021measure}. 
\begin{figure}[ht!]
    \centering
    \includegraphics[scale=.47]{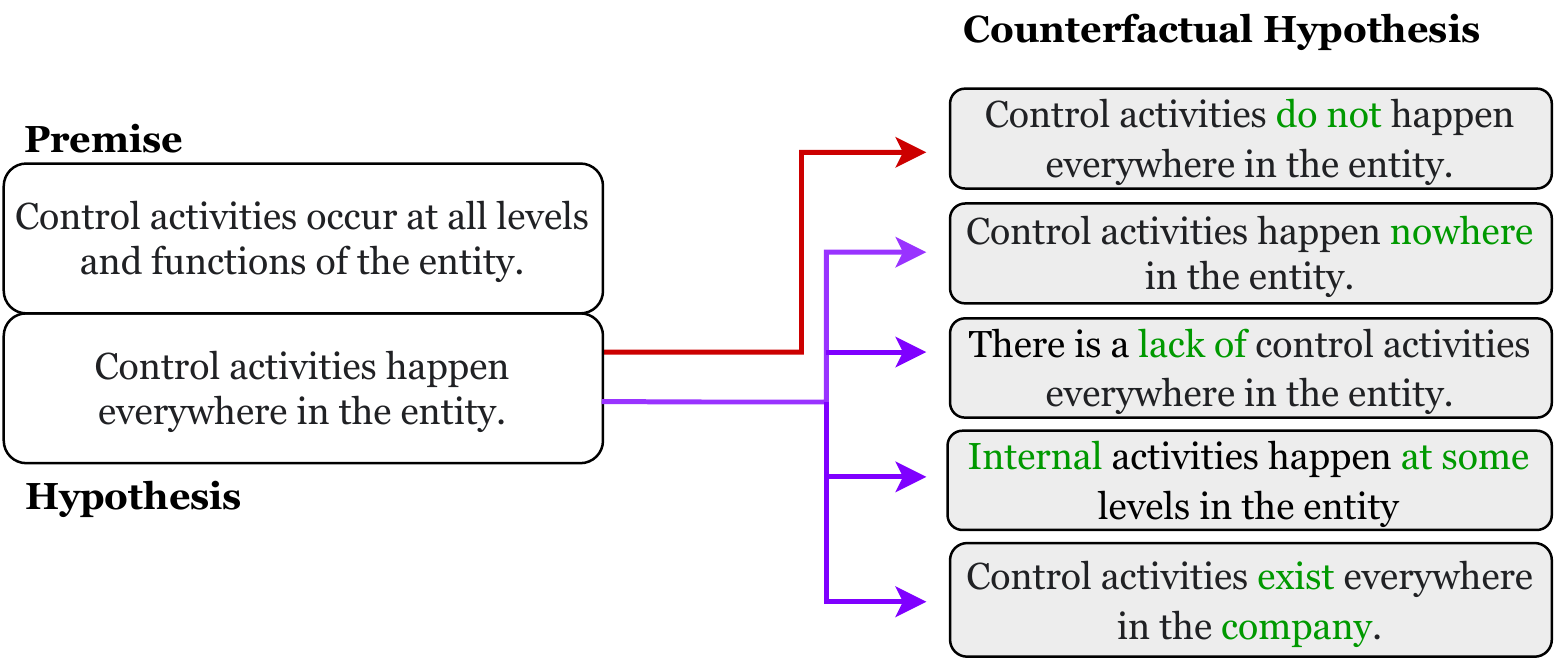}
    \caption{Diverse counterfactuals are generated for an MNLI example. The \textcolor{red}{red arrow} represents the most trivial way of generating a counterfactual hypothesis, while the \textcolor{viol}{violet arrows} depict several other perturbations that intervene on different predictive features.}
    \label{fig:nli_fig}
\end{figure}
Counterfactual data augmentation (CDA) has shown to improve model robustness to OOD data and input perturbations \citep{geva2021break,wu-etal-2021-polyjuice,paranjape-etal-2022-retrieval,khashabi-etal-2020-bang}. Alternate methods like debiasing data \cite{wu-etal-2022-generating} have also shown promising results on improving model robustness, but in this work we focus on CDA strategies. Recently, \citet{joshi-he-2022-investigation} find that diversity in the set of perturbations 
of different predictive features
 is key to the effectiveness of CDA (see Figure \ref{fig:nli_fig}). In this paper, we introduce \textbf{CO}unterfactual Generation via \textbf{R}etrieval and \textbf{E}diting (\textbf{CORE}) -- retrieval augmented generation framework for creating \emph{diverse} counterfactual perturbations. CORE combines dense retrieval with the few-shot learning capabilities of large language models, while using minimal supervision about perturbation type.



Retrieval-augmented  models 
\citep{guu2020realm,lewis2020retrieval}
 learn to search over a dense
 index of a text corpus to condition generation on retrieved texts and are especially 
effective at improving the diversity of generated text 
for paraphrase generation 
\citep{chen-etal-2019-controllable} and style-transfer \citep{xiao-etal-2021-transductive}.
CORE uses this insight by learning to retrieve \emph{counterfactual}  excerpts from a large text corpus.
Arbitrarily conditioning on these retrieved text excerpts 
to generate a rich set of counterfactual perturbations, without explicit supervision, can be challenging \citep{qin2022cold}. Instead, CORE uses few-shot prompting of massive pretrained language models, which is found to be effective at controlled generation tasks like arbitrary style-transfer \citep{reif-etal-2022-recipe}. CORE prompts GPT-3 \citep{brown2020language,wei2022chain} with \emph{a few} demonstrations of using these excerpts for counterfactual editing.

The CORE retriever  is a transformer-based bi-encoder model trained using contrastive learning \citep{le2020contrastive} on a small set of human-authored counterfactuals for the task. For each training example, CORE retrieves excerpts from an unlabeled  task-related corpus that bear a label-flipping counterfactual relationship with the original input instance. Retrieval may extract excerpts that have significant semantic drift from input text, while still containing relevant counterfactual phrases (Table \ref{tab:example_table}).
Using prompts, the CORE GPT-3 editor generates counterfactual edits to the input conditioned on the retrieved excerpts (and the original inputs). The prompts consist of instructions and a few demonstrations of using the retrieved text for editing.
Unlike prior work that use rule-based~\citep{ribeiro-etal-2020-beyond} or semantic  frameworks~\citep{wu-etal-2021-polyjuice, ross-etal-2022-tailor} and restrict perturbation types, CORE uses naturally occurring data to encourage perturbation diversity.

Intrinsic evaluation of CORE counterfactuals demonstrates a rich set of perturbation types which existing methods like \citet{wu-etal-2021-polyjuice} generate (Table \ref{tab:perturbation_coverage}) and  new perturbation types (Table \ref{tab:perturbation_diversity}) with more  diverse outputs (Table \ref{tab:IMDb_intrinsic}), without explicit supervision.
Our extensive data augmentation experiments and analyses show that the combination of retrieval and few-shot editing generates data for CAD that is effective in reducing model biases and improves performance on out of distribution (OOD) and challenge test sets. Perturbing only 3\% and 7\% of the data for NLI and Sentiment analysis respectively, we achieve improvements up to 4.5\% and 6.2\% over standard DA (Tables \ref{imdb_results},\ref{mnli_results}). 
 Additionally, we  show that CORE's learned retriever can assist humans in generating more diverse counterfactuals, spurring their creativity and reducing priming effects \citep{gardner-etal-2021-competency}.



\section{Related Work}

\paragraph{Counterfactual Data Augmentation}
There is growing interest in the area of CDA for model robustness, with early efforts focused on human-authored counterfactuals \citep{kaushik2020learning, gardner-etal-2020-evaluating}. However, manual rewrites can be costly and prone to systematic omissions.
Techniques have been proposed for the automatic generation of counterfactual data or contrast sets \citep{wu-etal-2021-polyjuice, ross-etal-2022-tailor, ross-etal-2021-explaining, bitton-etal-2021-automatic, asai2020logic, geva2021break, madaan2021generate, li-etal-2020-linguistically}.
Existing techniques rely on using rules/heuristics for perturbing sentences 
\citep{webster2020measuring, dua-etal-2021-learning, ribeiro-etal-2020-beyond, asai2020logic}, or using  sentence-level semantic representations (eg. SRL) and a finite set of structured control codes 
\citep{geva2021break, ross-etal-2022-tailor, wu-etal-2021-polyjuice}.
However, \citet{joshi-he-2022-investigation} find that a limited set of perturbation types further exacerbates biases, resulting in poor generalization to unseen perturbation types. Generally, creating an assorted set of \emph{instance-specific} perturbations is challenging, often requiring external knowledge \citep{paranjape-etal-2022-retrieval}.
\\\noindent\textbf{Retrieval Augmented Generation}
Retrieving task-relevant knowledge from a large corpus of unstructured and unlabeled text has proven to be very effective for knowledge-intensive language generation tasks like question answering \cite{lewis2020retrieval}, machine translation \citep{gu2018search} and dialogue generation \citep{weston-etal-2018-retrieve}. 
Retrieval has also been used for paraphrase generation \citep{chen-etal-2019-controllable} and style-transfer \citep{xiao-etal-2021-transductive} to specifically address the lack of diversity in generations from pretrained language models. In a similar vein, CORE uses learned retrieval for counterfactual generation. While \citet{paranjape-etal-2022-retrieval} use off-the-shelf retrieval models to generate counterfactuals for QA, learning to retrieve counterfactuals is non-trivial for problems other than QA. CORE provides a recipe to train retrieval for general tasks. \\\noindent\textbf{In-context learning}
Massive language models like GPT-3 have been found to be  effective at controlled generation tasks like arbitrary style-transfer  \citep{reif-etal-2022-recipe}, counterfactual reasoning \citep{frohberg-binder-2022-crass}, step-wise reasoning for complex problems \citep{wei2022chain,zhou2022least}, and dataset generation \citep{liu2022wanli}, by learning \emph{in-context} from few-shot demonstrations and natural language instructions \citep{wei2021finetuned}. While GPT-3 has been used for data augmentation, it has not been used for counterfactual generation, which is fundamentally different in nature.


\begin{figure*}[t!]
    \centering
    \includegraphics[scale=.5]{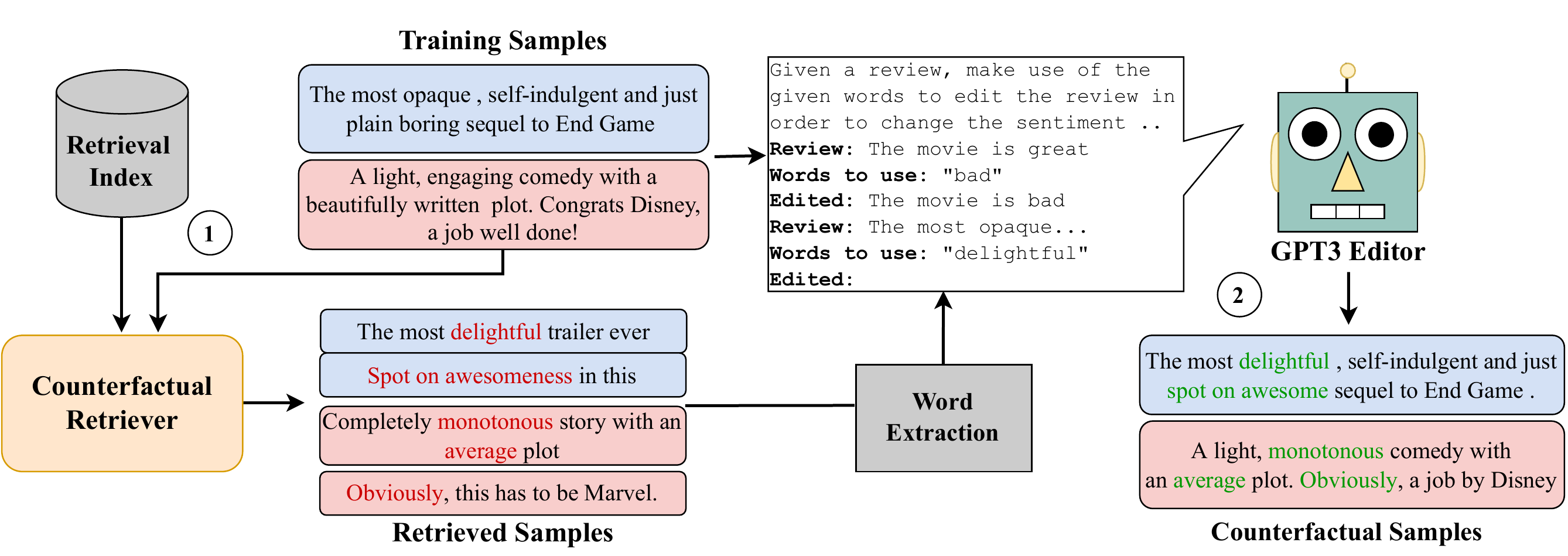}
  \caption{Overview of CORE: \textbf{CO}unterfactual \textbf{R}etrieval \textbf{E}diting framework. With the help of the \circled{1} trained counterfactual retriever we retrieve text excerpts from a large text corpus. These text excerpts are passed through a simple word extraction module that extracts all the non stopwords, which are then used by \circled{2} the Editor to edit the given training instances to generate minimally edited label flipped instances.  }
    \label{fig:teaser}
\end{figure*}

\section{Method}

A high level overview of CORE is shown in Figure~\ref{fig:teaser}. The first stage (\S\ref{sec:cfdpr}) retrieves counterfactual excerpts from a large unlabeled corpus related to the target task. In the second stage (\S\ref{sec:editor}), retrieved excerpts are supplied, along with instructions and demonstrations, as a language-modeling prompt to GPT-3 to counterfactually edit input text. The resultant data is used for augmentation in \S\ref{results:augmentation}.
We describe each stage below;
additional implementation details are provided in
Appendix \ref{appendix:implimentation_details}.

\subsection{CF-DPR: Counterfactual Dense Passage Retriever} \label{sec:cfdpr}


Our counterfactual retriever is based on the dense passage retrieval (DPR) framework \citep{karpukhin-etal-2020-dense}. CF-DPR  retrieves similar instances from a large unlabeled corpus that have different labels.
Formally, given a training set, $N(x)=\{(x_{1},y_{1}), (x_{2}, y_{2}), \dots , (x_{n}, y_{n})\}$, for a text classification task and a large corpus $S$, 
CF-DPR retrieves samples $C(x) = \{\hat{x}_{1}, \hat{x}_{2}, \dots, \hat{x}_{n}\}$  from $S$ such that the associated labels for samples in $C(x)$ are not the same as the \emph{corresponding} labels in $N(x)$. Specifically, $\hat{y}_{i} \neq y_{i} \forall i \in (0,n)$, where $\hat{y}_i$ is the class label for retrieved sample $\hat{x}_i$. In Figure \ref{fig:teaser}, for the input ``The most opaque , self-indulgent and just plain boring sequel to End Game.'', CF-DPR retrieves the excerpts ``The most delightful trailer ever" and ``Spot on awesomeness in this''.

\noindent \textbf{Training} We use the same contrastive learning objective as \citet{karpukhin-etal-2020-dense} to train the bi-encoder model. It consists of two independent BERT \citep{devlin2018bert} encoders: a query encoder $P$ that encodes $x_i$ in $N(x)$ as $p_i$ and a document encoder $Q$ that encodes text excerpts in $S$ as $q_i$.
To train the bi-encoder, we  use a small seed training dataset $[q_{i}, p^{+}_{i}, p^{-}_{i}]_{i=1}^m$ of size $m$ containing $m < |N(x)|$ positive and negative retrieval samples.
For a given training instance $q_{i}$, we use its corresponding positive sample $p^{+}_{i}$ and hard negatives $p^{-}_{i}$'s to optimize the following loss function.
\begin{dmath}
    L(\{q_{i}, p^{+}_{i}, p^{-}_{i,1}, .. p^{-}_{i,n}\}_{i=1}^{m}) =
    -\log\frac{e^{sim(q_{i}, p^{+}_{i}})}{e^{sim(q_{i}, p^{+}_{i})} +  \sum_{j=1}^{n} e^{sim(q_{i}, p^{-}_{i,j})}}
    \label{eq:cfdpr_loss}
\end{dmath}
To model the task of counterfactual retrieval, for each training instance $x_{i} = q_{i}$, we use the corresponding counterfactual instance as the positive sample ($p^{+}_{i}$) and use paraphrases of $q_{i}$ as the hard negative ($p^{-}_{i}$). Positive samples can be obtained from a seed dataset consisting of manually authored counterfactuals for existing NLU datasets like IMDb, and SNLI \citep{kaushik2020learning,gardner-etal-2020-evaluating}. This manual data is of the form $T = \{(q_1, p^{+}_{1}), (q_2, p^{+}_{2}), \dots, (q_m, p^{+}_{m})\}$.
We make use of the diverse paraphraser model \citep{krishna-etal-2020-reformulating} that generates paraphrases as hard negatives for  $\{q_1, q_2, \dots, q_m\}$, $\{p^{-}_{1}, p^{-}_{2}, \dots, p^{-}_{m}\}$. Contrastive training pulls counterfactual samples $p^{+}_{i}$ closer to $q_i$ and pushes semantically identical sentences $p^{-}_{i}$ away from $q_i$. We show that this counterfactual retrieval framework can be used to retrieve counterfactuals for tasks with only a small amount of seed training data (\S \ref{sec:method_sentiment}). 
Additional details about training and evaluation of the trained CF-DPR are in Appendix~\ref{appendix:implimentation_details}.


\paragraph{Inference} We create $S$ for a specific task dataset using (1) text corpora that have similar domains as that dataset and (2) other datasets for the same task. For instance, for sentiment analysis over IMDb, we use a large (1.6 million) corpus of movie reviews from Amazon \citep{10.1145/2507157.2507163} and  the Yelp review dataset \citep{asghar2016yelp}.
We encode the entire search corpus using the trained document encoder and index it using FAISS  \citep{johnson2019billion}. For every input training instance $x_i$, we retrieve top \textit{k} relevant counterfactuals $\{\hat{x}^1_i, \hat{x}^2_i, \dots, \hat{x}^k_i\}$.  We refer to these as \textit{CF-DPR counterfactuals}.

\subsection{GPT-3 Editor}
\label{sec:editor}

The retrieved counterfactuals often contain relevant phrases that perturb predictive features in different ways (``opaque'' $\rightarrow$ ``delighful'', ``boring'' $\rightarrow$ ``spot-on awesome'' in Figure~\ref{fig:teaser}), but are typically not a minimally edited version of the training sample. ``The most delightful trailer ever'' has the opposite sentiment as the original review, but is about another entity.
To incorporate perturbation diversity while controlling for minimality, CORE uses an editor module.
The Editor takes the training sample and retrieved counterfactuals as input and generates a minimally edited counterfactual. This involves selecting parts of the retrieved text that maybe useful in flipping the original text's label and then seamlessly integrating them into the original input. This fine-grained instance-specific style-transfer task can be hard to find supervision for. 

We use GPT-3~\citep{brown2020language} 
since it has been successfully used for fine-grained style transfer with few-shot supervision in prior work~\citet{reif2021recipe}. For instance, GPT-3 can learn from natural language constraints in its prompt, such as ``include the word balloon,'' for constrained decoding. 
To prompt GPT-3, we make use of the set of instructions that were provided in \citet{kaushik2020learning} to instruct crowd workers to author counterfactuals. We also append four human authored demonstrations of incorporating retrieved data, depicting various perturbation types. 


Following \citet{reif-etal-2022-recipe}, we simplify our demonstrations by extracting keywords from the retrieved samples and providing them as token-level constraints in the prompt.
To encourage the model to perturb certain classes of words, we remove determiners, conjunctions, and punctuation from the retrieved samples and tokenize the rest of the input into a list of keywords: [$w_1, \dots, w_n$]. 
The resultant demonstration in our prompt thus becomes: ``Input: ... Words to use: [$w_1, \dots, w_n$], Edited:...''.  This is motivated by 
\citet{wu-etal-2021-polyjuice}'s observation that  perturbing certain classes of words (like preposition and adjectives) leads to better counterfactual generation. More details about the prompt construction are in Appendix \ref{appendix:implimentation_details}. 



\section{Experimental Setup}
\label{sec:exp_setup}
We generate CORE counterfactuals for two tasks -- sentiment classification of movie reviews and natural language inference. We describe task-specific details about CORE training and inference below. 
\begin{table*}[t]
    \centering
    \small
    \begin{tabularx}{\linewidth}{p{0.23\linewidth}|p{0.36\linewidth}|p{0.34\linewidth}}
    \toprule
    \multicolumn{3}{c}{\textbf{IMDb}} \\
    \midrule
    \multicolumn{3}{p{0.9\textwidth}}{\textbf{Review:} It's the worst movie I've ever seen. The action is so unclear, work of cameras is so poor, actors are so affected ... and this lamentable 5 minutes of Arnie on the screen. My advice from the bottom of my heart - don't watch it unless you like such a low class torture. \textbf{Sentiment:} Negative}\\
    \midrule
    \bf CF-DPR & \bf GPT-3 & \bf CORE  \\
    Probably the dozenth time I've seen it. This is another such effort. This is one spectacular production. that is definitely worth taking advantage of. 

    & It's one of the \textcolor{blue}{best} movies I've ever seen. The action is so \textcolor{blue}{clear}, work of cameras is so \textcolor{blue}{great}, actors are so \textcolor{blue}{unaffected} ... and this \textcolor{blue}{wonderful} 5 minutes of Arnie on the screen. My advice from the bottom of my heart - watch it unless you don't like such a \textcolor{blue}{high class} torture.
    & It's the \textcolor{blue}{best} movie I've ever seen. The action is so \textcolor{blue}{clear}, work of cameras is so \textcolor{blue}{great}, actors are so \textcolor{teal}{professional} ... and this \textcolor{teal}{spectacular} 5 minutes of Arnie on the screen. My advice from the bottom of my heart - \textcolor{teal}{definitely} watch it if you like such a \textcolor{blue}{high class} \textcolor{teal}{production}.
    \end{tabularx}
    \begin{tabularx}{\linewidth}{p{0.40\linewidth}|p{0.25\linewidth}|p{0.25\linewidth}}
    \toprule
    \multicolumn{3}{c}{\textbf{~~~ MNLI}} \\
    \midrule
    \multicolumn{3}{l}{\textbf{Premise}: and my my uh taxes are a hundred and thirty five. \textbf{Hypothesis}: My taxes are $\$135$ \textbf{(Entails)}} \\
    \textbf{CF-DPR} & \textbf{GPT-3} & \textbf{CORE} \\
    My sister spent over $\$2,000$ on a computer that she'll probably never use. &  My taxes are \textcolor{blue}{not} $\$135$. &  My taxes are \textcolor{teal}{probably over} $\$135$ \\
    \midrule
    \multicolumn{3}{l}{\textbf{Premise}:After the battle against the slave lord, the group needed it. \textbf{Hypothesis}:The slaves battled the group.}\textbf{(Contradicts)}\\ 
    \textbf{CF-DPR} & \textbf{GPT-3} & \textbf{CORE} \\
     The slaves never fought against their tyrannical captors. & The slaves battled the \textcolor{blue}{slave lord}. &  The slaves \textcolor{teal}{fought} against their \textcolor{teal}{captors}.\\
    
     \bottomrule
    \end{tabularx}
    \caption{CORE generated counterfactuals for IMDb and MNLI, along with the CF-DPR retrieved outputs and the independent GPT-3 Editor. For both the tasks, the retriever introduces several \textcolor{teal}{new words/phrases} in the outputs.}
    \label{tab:example_table}
\end{table*}

\subsection{Sentiment Classification} 
\label{sec:method_sentiment}
\paragraph{Task Dataset ($N(x)$)} We create CORE counterfactuals for the IMDb movie review dataset \citep{maas-etal-2011-learning}, which has been used to manually create contrastive data \citep{kaushik2020learning, gardner-etal-2020-evaluating}. This dataset presents unique challenges due to the longer average length of reviews (233 words), that existing counterfactual generation techniques \citep{wu-etal-2021-polyjuice} struggle at.

\noindent\textbf{CF-DPR training data} ($p^{+}_{i}, p^{-}_{i}$) \citet{kaushik2020learning} augment a subset of the IMDb dataset (1.7K examples) with human edited counterfactuals, which we use to train CF-DPR. 
Negative pairs  $p^{-}_{i}$ are created by paraphrase models.

\noindent\textbf{Task-specific corpus ($S$)} We use datasets that are of similar domain --- Amazon Movie reviews \citep{10.1145/2507157.2507163}, Yelp reviews \citep{asghar2016yelp}, and IMDb reviews \citep{maas-etal-2011-learning}. Our initial experiments indicated that indexing full movie reviews did not yield good CF-DPR performance,
owing to more dense retrieval noise when encoding longer contexts \citep{luan2021sparse}.
Hence, we sentence tokenize the reviews and index each sentence independently. The search corpus contains approximately 8 million sentences.

\noindent\textbf{Subset to augment $N(x)$} We generate CORE counterfactuals for the same subset of $1.7k$ reviews from IMDb chosen by \citet{kaushik2020learning} in order to make fair comparisons in \S \ref{results}.  

\noindent\textbf{Key-word list $[w_1, \dots, w_n]$} We observed that several sentences in the review contain descriptive information (plot, cast, etc) that don't convey sentiment. In \citet{kaushik2020learning} too, humans edited a few sentences to change the sentiment of a long review. We extract these sentences containing polarity features, use them as queries to CF-DPR,  and retrieve their counterfactuals. 
This selection strategy can also be extended to any review by learning a sentence selection model on annotated data   \citep{deyoung-etal-2020-eraser}. 
The keyword list 
is extracted from retrieved outputs as described in \S\ref{sec:editor}.
Note that we do not impose any restrictions on the editor regarding which sentences to edit. 

\subsection{Natural Language Inference}
\label{sec:method_nli}

\paragraph{Task Dataset $N(x)$} We focus on MNLI \citep{N18-1101}, a popular NLI dataset that tests for complex language reasoning.

\noindent \textbf{CF-DPR training data $p^{+}_{i}, p^{-}_{i}$} 
We use the inherent paired nature of MNLI. In MNLI, given a premise, annotators are asked to manually write one sentences that entail, contradict or are neutral to the premise. These three hypotheses serve as mutual counterfactuals. 
In this work, we limit counterfactual perturbations to entailment(E)$\rightarrow$contradiction(C) and vice-versa, to simplify the different permutations of positives and negatives required for CF-DPR training. We find that including the neutral class leads to increasingly noisy retrieved data, as the semantic differences between neutral class and the other two NLI classes are subtle and hard to distinguish. 
In Equation ~\ref{eq:cfdpr_loss},  $q_{i}$ is generated by concatenating the premise and hypothesis separated by the special token \texttt{[SEP]}. 
For every such input, $p^{+}_{i}$ is a hypothesis from the counterfactual class, while $p^{-}_{i}$ are diverse paraphrases of the original hypothesis.
\\\noindent\textbf{Task-specific corpus ($S$)}
is constructed by combining the following NLI datasets \citep{N18-1101, snli:emnlp2015, liu2022wanli} in addition to the source corpus
\footnote{\href{http://www.anc.org/}{http://www.anc.org/}} 
that was used to generate premises in MNLI. We also include tokenized wikipedia articles \citep{merity2016pointer} as several domains in MNLI (Eg. travel, government) are related. The search corpus contains approximately 7 million text excerpts.
\\\noindent \textbf{Subset to augment $N(x)$} To compare with the state-of-the-art data augmentation technique for MNLI, WaNLI \citep{liu2022wanli}, we choose a subset of the MNLI dataset for augmentation based on their selection strategy. WaNLI uses dataset cartography \citep{swayamdipta-etal-2020-dataset} to select the most ambiguous examples --- where model confidence across training epochs is low and variance is high --- for augmentation. We generate 9.5K additional examples in two classes.
\\\noindent \textbf{Cross-Encoder}
We incorporate a re-ranker module to boost retrieval results for MNLI, that uses a cross-encoder architecture \citep{thakur-etal-2021-augmented} to jointly encode query $q_i$ and top-K documents retrieved by the bi-encoder. Given a $q_i$ and $K$ retrieved sentences from the bi-encoder, the re-ranker learns to classify them as positive or negative. During inference, bi-encoder outputs are re-ranked based on their cross-encoder probability.
The cross encoder is trained on the binary classification task on the same seed dataset as the bi-encoder. Retrieval performances are reported in Appendix \ref{appendix:additional_results}.







\begin{table*}[t]
    \centering
    \small
    \begin{tabular}{l|g|k|k|k|b|b}
        \toprule
        \bf Train $\downarrow$ \bf Test $\rightarrow$ & \bf \small IMDb & \bf \small Senti140 & \bf \small SST2 & \bf \small Yelp & \bf \small IMDb Cont  & \bf \small IMDb CAD \\
        \midrule
        IMDb  &                                    
        90.98 &
 75.30 & 
84.63 &
90.04 &
81.35 &
83.76
\\
        ~~~~~ + CAD (Human, Clean)   &  
        91.3  & 
75.77 & 
88.19 & 
91.31 & 
\bf 86.68  & 
\bf 88.54
         \\
         \midrule
        ~~~~~ + Data Augmentation&
    91.23  &
69.67 &
73.16 &
84.48 &
82.17 &
84.44

         \\
                 \midrule
        ~~~~~ + GPT-3 counterfactuals
& 91.1  &
 75.90 &
88.41 &
92.31 &
83.40 &
86.22

         \\
        ~~~~~ + CFDPR counterfactuals  
& 91.51  &
74.09  &
88.30  &
91.19  &
 79.30  &
80.90
         \\
        
        ~~~~~ + CORE counterfactuals
 & 91.18 &
\bf 78.12 &
\bf 90.82 &
\bf 92.01 &
84.63 &
86.35

         \\
        
        \bottomrule
    \end{tabular}
    \caption{Accuracies of various data augmentation strategies on IMDb  (Section~\ref{results:augmentation}). CAD augmentation is the noise free involving human intervention while the rest are noisy.
    Although \colorbox{grey}{in-domain} performance is unaffected, we can see notable gains on all \colorbox{green}{out-of-distribution datasets} \citep{go2009twitter, socher-etal-2013-recursive, asghar2016yelp} and also competative gains on \colorbox{LightCyan}{contrast} \citep{gardner-etal-2020-evaluating} and \colorbox{LightCyan}{CAD} \cite{kaushik-etal-2021-efficacy} test sets. Statistical variance in results across runs is < 0.5 points.}
    \label{imdb_results}
\end{table*}

\begin{table*}[t]
    \centering
    \resizebox{1\textwidth}{!}{
    \begin{tabular}{l | g | k | k | k | b | b |  b|  b b  b}
    \toprule
      \multirow{2}{*}{\bf Train $\downarrow$ \bf Test $\rightarrow$} & \cellcolor[gray]{0.9}  & 
      & 
      & 
      & 
       & 
    &  
      \bf NLI-Adv & \multicolumn{3}{| b}{\bf ANLI} \\
      
     & \multirow{-2}{*}{\cellcolor[gray]{.9}\bf MNLI} 
     & \multirow{-2}{*}{\bf QNLI}
     & \multirow{-2}{*}{\bf SNLI} 
     & \multirow{-2}{*}{\bf WaNLI}
     & \multirow{-2}{*}{\bf HANS}
     & \multirow{-2}{*}{\bf Diagno}
     & \bf \small LI_LI & \bf R1 & \bf R2 & \bf R3 \\
    \midrule
    MNLI 
    & 87.66 
    & 50.57
    & 83.82
    & 59.10
    & 68.22
    & 61.11
    & 90.39
    & 32
    & 30
    & 28.58\\
~~~~~  + WaNLI (Human, Clean)
 & 88.02 
 & 50.57
 & 84.85
 & 58.56
 & 70.90
 & 62.59
 & 91
 & 34
 & 29.40
 & 30.25
 \\
~~~~~  + Tailor 
 & 88.28 
 & 50.53
 & 83.03
 & 60.66
 & 70.73
 & 62.59
 & 90.25
 & 34.20
 & 29.60
 & 29.08
 \\
\midrule
~~~~~  + GPT3 counterfactuals 
                                               
& 87.73
& 50.70
& 82.74
& 58.96
& 64.43                                                     
& 61.50
& 88.07                              
& 32.80                               
& 29.20 
& 32.80
\\
~~~~~  + CFDPR counterfactuals                              
&  87.79
& 44.99
&  83.06   
&  59.14                                                      
&  66.49                                              
& 62.13                                                       
& 89.70 
& 33.50 
& 29.70                                    
& 29.50
\\
~~~~~ + CORE counterfactuals 
& 87.97
& 50.52
& 84.34
& \bf 60.80
& \bf 72.57
& 62.32
& 90.98
& 33 
& 28.90
& \bf 30.58
\\
~~~~~ + WaNLI + CORE counterfactuals 
&  88.31
& 50.61
 & 85.03
 & 58.96
 & 70.55
 & 62.50
& 90.31
& \bf 34.90
& 29
& 30.25 \\

    \bottomrule

    \end{tabular}
    }
    \caption{Accuracies of data augmentation for CORE and baselines on MNLI (Section~\ref{results:augmentation}). 
    CORE is competitive (within variance) or improves over WaNLI and MNLI baseline in almost all cases. We have competitive performance on both,  \colorbox{green}{out-of-distribution datasets} \citep{rajpurkar2016squad,bowman-etal-2015-large,liu2022wanli}, \colorbox{LightCyan}{challenge-sets} \citep{mccoy2019right, wang-etal-2018-glue, naik-etal-2018-stress, glockner-etal-2018-breaking} and \colorbox{LightCyan}{Adversarial NLI (ANLI)} \cite{nie-etal-2020-adversarial}. Statistical variance in results across runs < 0.5 points.}
    
    \label{mnli_results}
\end{table*}


\section{Experimental Results}
\label{results}
We first list the various augmentation strategies against which we compare CORE (\S\ref{sec:aug_strat}) followed by data augmentation results 
(\S\ref{results:augmentation}). In (\S\ref{sec:ablation}) we highlight the need for a counterfactual retriever, and in (\S\ref{results:instrinsic_eval}) we intrinsically evaluate CORE counterfactuals on their quality and perturbation diversity.

\subsection{Counterfactual Data Augmentation}
\label{sec:cad}
We augment the full training datasets with $|N(x)|$ CORE counterfactuals. We fine-tune DeBERTa base model \citep{he2021deberta} on the  combined dataset. 

\subsubsection{Augmentation Strategies}
\label{sec:aug_strat}
We compare augmentation with CORE to strong DA baselines and ablations to different parts of our data generation pipeline. For all the strategies we augment with the same number of instances.

\paragraph{Data Augmentation Baselines}
In order to evaluate the effectiveness of CDA over adding more in-domain data,
we add the same number of in-domain training examples. For sentiment classification, we add $1.7k$ movie reviews randomly sampled from the Amazon Movie reviews dataset \citep{10.1145/2507157.2507163}.
For MNLI, we use the \textbf{WaNLI} \citep{liu2022wanli} dataset constructed using MNLI, that showed impressive OOD gains. We randomly sample 9K reviews from E and C classes in WaNLI.
\paragraph{Counterfactual Data Augmentation} 
For MNLI, we compare against \textbf{Tailor} \citep{ross-etal-2022-tailor} using the \texttt{SWAP\_CORE} control code as it results in label flipping perturbations. We do not compare with Polyjuice \citep{wu-etal-2021-polyjuice} as it requires complete human relabelling hence it would not be a fair comparison. \\
\noindent \textbf{Human generated data} For IMDb, we also compare against human-authored counterfactual data, \textbf{CAD} \citep{kaushik2020learning}.
Amazon reviews, WaNLI, and CAD \emph{involve human supervision} for dataset construction and are generally noise-free, unlike CORE data (see \S \ref{results:label_correctness} for noise estimates).

\noindent \textbf{GPT-3 Counterfactuals} To ablate the effect of conditioning on retrieval, the \textbf{GPT-3 counterfactual baseline} edits inputs into counterfactuals based only on a prompt consisting of task instruction and demonstrations with NO keyword lists. 

\noindent \textbf{CF-DPR Counterfactuals}
To ablate the GPT-3 few-shot editor in the CORE pipeline, the \textbf{CF-DPR counterfactual baseline} that uses retrieved outputs of CF-DPR as counterfactuals for augmentation and does no few-shot editing. See Appendix~\ref{appendix:implimentation_details} for more task-specific  details of this baseline.



\subsubsection{Results}
\label{results:augmentation}


\paragraph{IMDb}
Table~\ref{imdb_results} shows that using just $1.7k$ (7\%) human-annotated contrastive data to train the CF-DPR model, 
followed by the GPT-3 editing step, CORE obtains a performance gain of up to 6.2\% over the IMDb un-augmented baseline. We compare against human-authored \citet{kaushik2020learning} (CAD) \textbf{clean} examples in the CAD dataset and MiCE, an automatic counterfactual editing technique. We find that CORE is especially effective at OOD improvements on the Senti140, SST2 and Yelp datasets (by 3.66\% points on average); despite augmenting data with noisy labels and with no explicit human supervision for editing. We hypothesize that this may be because of priming biases in human-authored counterfactual data \citep{bartolo2021models} and more diversity in CORE counterfactuals. A more detailed analysis is 
presented in  \S\ref{results:instrinsic_eval}. 

Standard in-domain augmentation (i.e. bearing no counterfactual relationship with original data) is not as effective at improving performance on OOD and contrastive sets (Table~\ref{imdb_results},Table~\ref{mnli_results}), thus highlighting the importance of counterfactualy generated data.

Ablating individual parts of the pipeline, i.e GPT-3 based editing and CF-DPR retrieval -- we observe that individual components are less effective than the combination of both techniques in CORE. CF-DPR retrieval may create reviews that incoherent and have a large semantic shift from the original review. Though the independent GPT-3 editor generates reviews that are minimally edited, they may contain recurring perturbation types (Table~\ref{tab:example_table}), limiting its efficacy for CDA. Combining the two helps overcome individual drawbacks.

\paragraph{MNLI}
Using the DeBERTa-base model, with just augmenting $9.5k$ (3\%) of the data we get improvements of up to 4\% over the unaugmented dataset (Table~\ref{mnli_results}). The improvements are particularly on the WaNLI evaluation set and adversarially designed test sets HANS and ANLI. Compared to Tailor, CORE achieves a 2\% and 1.3\% gain on HANS and SNLI, respectively. Once again, retrieval and GPT-3 editing are not as effective individually as the combination of the two in CORE. 

Independent GPT-3 generated counterfactuals are biased towards simple perturbations (\S\ref{results:diversity}) and augmenting with these counterfactuals hurts performance on HANS and on  Lexical Inference test set (LI\_LI). CORE is also competitive with WaNLI data augmentation, which is noise-free in-domain data constructed with the same subset selection strategy as used by CORE. 
We also consider augmenting with both CORE and WaNLI counterfactuals, which results in orthogonal benefits and improvements on SNLI and ANLI datasets. 




\begin{figure}[t]
    \centering
    \includegraphics[scale = 0.27]{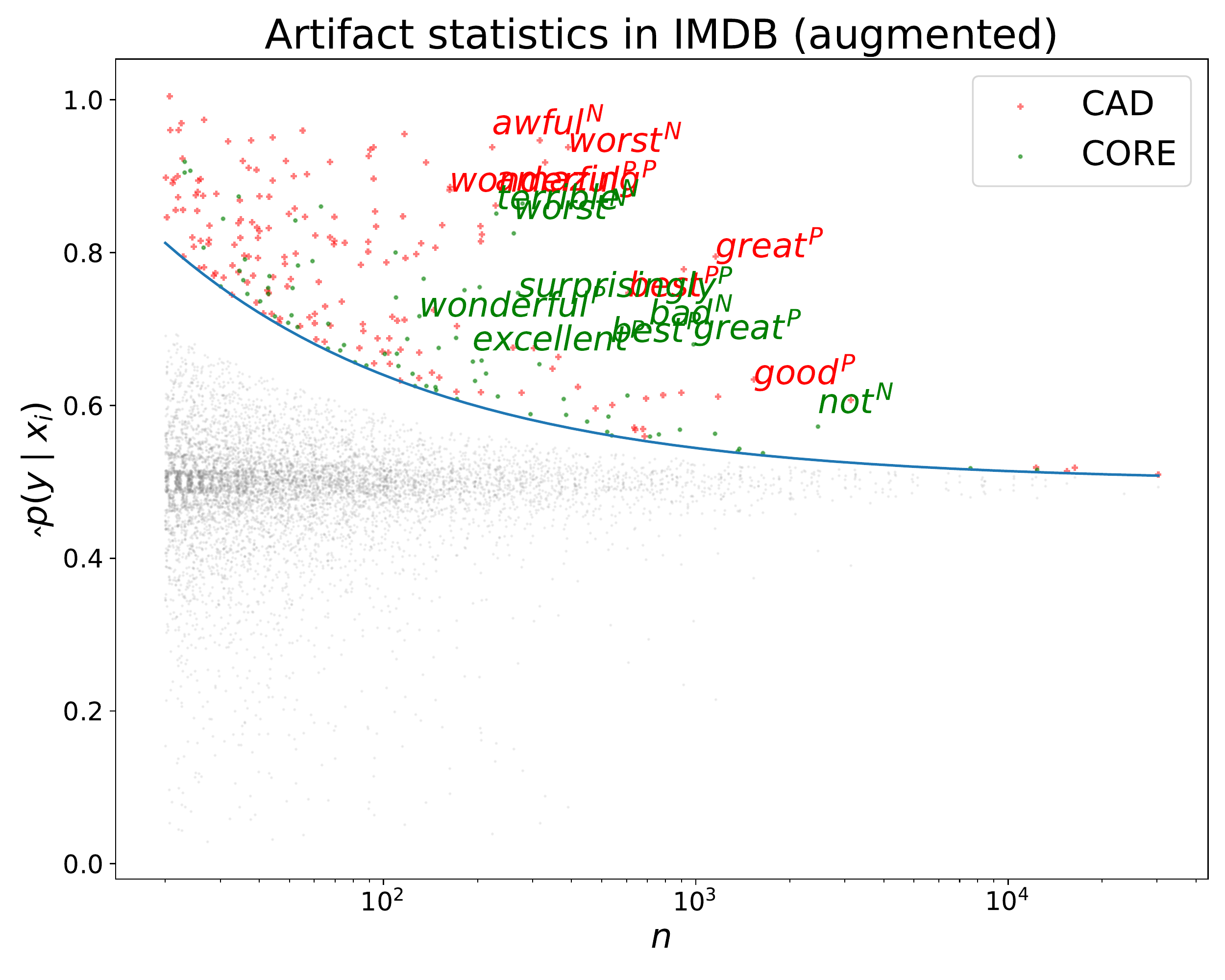}
    \caption{
    A statistical test for deviation from a competency problem \citep{gardner-etal-2021-competency}, where no word should be informative about class label. Words above the blue line  have detectable correlations with labels. CAD displays more statistical bias compared to CORE.}
    \label{fig:z_stats}
\end{figure}

\begin{table}[t]
    \centering
    \resizebox{0.5\textwidth}{!}{
    \begin{tabular}{l | c | c | c | c}
    \toprule
      \multirow{2}{*}{\bf Train $\downarrow$ \bf Test $\rightarrow$} & 
      & 
      & 
    &  
      \bf NLI-Adv 
    \\
      
    & \multirow{-2}{*}{ \bf MNLI}
     & \multirow{-2}{*}{\bf HANS}
     & \multirow{-2}{*}{\bf Diagno}
     & \bf \small LI_LI 
     \\
    \midrule
    MNLI 
    & 87.66 
    & 68.22
    & 61.11
    & 90.39
    \\
~~~~~ + TF-IDF + CORE Editor 
& 87.89
& 60.51
&  61.77
& 87.49
\\
~~~~~ + CFDPR + CORE Editor 
& 87.97
& \bf 72.57
& \bf 62.32
& 90.98
\\

    \bottomrule

    \end{tabular}}
    \caption{Comparison between CFDPR and TF-IDF based retriever. We observe that using a general text retriever hurts performance.}
    
    \label{main_ablation_mnli_results}
\end{table}

\begin{table*}[ht!]
\centering
\small
\begin{tabular}{p{0.40\linewidth}|p{0.52\linewidth}}
\toprule
\bf Polyjuice \citep{wu-etal-2021-polyjuice} & \bf CORE\\
\midrule 
\multicolumn{2}{l}{1. The problem with all of this : It 's \textbf{not} really funny.}\\
1. [delete] The problem with all of this: It is \textcolor{red}{\st{not}} very funny. & 1. The \textcolor{blue}{great} thing about all of this: It's \textcolor{blue}{funny, considered anyway - for no particular reason.}\\
\multicolumn{2}{l}{2.  Munch 's screenplay is \textbf{tenderly} observant of his characters .} \\
2. [insert]and Munch's screenplay is \textcolor{red}{not} observant of his characters. & 2. Seldom does a \textcolor{blue}{contrived screenplay comes across} that is ever so observant of its characters.\\
\midrule
\multicolumn{2}{l}{1. As a film director , LaBute continues to \textbf{improve} .} \\
1. [lex] As director, LaBute continues to \textcolor{red}{crap}. & 1. As a film director, LaBute is \textcolor{blue}{not as good as one might hope.}\\
\multicolumn{2}{l}{2. The movie is a \textbf{mess} from start to finish .} \\
2. [lex] It is a \textcolor{red}{wonderful} movie from start to finish. & 2. I went to the movie weeks ago and \textcolor{blue}{enjoyed it from start to finish.}\\
\bottomrule
\end{tabular}
\caption{Examples of perturbations on SST2 data. Polyjuice often uses restricted patterns like perturbing ``not'' or using \emph{antonyms},  unlike CORE's non-trivial perturbations conditioned on retrieved data.}
\label{tab:perturbation_diversity}
\end{table*}

\subsection{Retrieval Ablation}
\label{sec:ablation}
To understand the importance of a trained counterfactual retriever, we ablate the retrieval stage of CORE by using a simple TF-IDF based retriever. We use the same GPT-3 Editor to generate the counterfactuals, only replace the CFDPR counterfactual retriever with a TF-IDF retriever. The TF-IDF based CORE counterfactuals exacerbate biases as seen in Table~\ref{main_ablation_mnli_results}, significantly hurting performance by 8\% and 3\% on HANS and Adv. NLI, full table in Appendix (Table~\ref{ablation_mnli_results}). This highlights the need for a counterfactual retrieval module that retrieves counterfactual words/phrases in an unconstrained manner without limiting potential semantic edits.

\subsection{Intrinsic Evaluation}
\label{results:instrinsic_eval}
To analyze the source of its empirical gains of CORE in \S \ref{results:augmentation}, we evaluate the label correctness, closeness, and perturbation diversity.

\paragraph{Label Correctness}
\label{results:label_correctness}
Unlike Polyjuice \citep{wu-etal-2021-polyjuice} CORE encourages label flipping behavior during the generation process, \emph{possibly} at the cost of label correctness. 
We quantify label correctness of the generated data by manually annotating a sample of 100 data points for IMDb and MNLI. Our analysis show that noise levels are 41\% for IMDb and 40\% for MNLI. 
CORE's augmentation benefits (\S \ref{results:augmentation}) persist even when compared to noise-free data (CAD for IMDb and WaNLI for MNLI), underscoring the importance of diversity.

\paragraph{Closeness and Diversity}
\label{results:diversity}
The intuition behind CORE is that effective CDA requires perturbation of various kinds so that perturbation bias is not exacerbated \citep{joshi-he-2022-investigation}. To measure this effect, on the IMDb dataset, we compare with CAD and another prior work on CAD generation, MiCE \citep{ross-etal-2021-explaining}. In Figure~\ref{fig:z_stats}, for the augmented subset we plot the probability of a token predicting class labels as a function of token count. 
We observe that CAD \citep{kaushik2020learning} has several outliers tokens that have strong bias towards a label, compared to CORE.
\par Since counterfactuals are meant to be \textit{minimally} edited instances of the original input \cite{gardner-etal-2020-evaluating}, we analyse the closeness of the generated counterfactuals to the original text. To check how close these generated counterfactuals are to the original instance, we measure the Levenshtein edit distance between the two. To quantify diversity, we also measure self-BLEU \citep{10.1145/3209978.3210080}. The self-BLEU score is computed between edited counterfactual and the original text on the IMDb dataset. Since self-BLEU and Levenshtein are opposite in nature, we ideally want the counterfactuals to strike a balance between the two \citep{wu-etal-2021-polyjuice}. As shown in in Table~\ref{tab:IMDb_intrinsic}, CORE counterfactual are more diverse (lower Self-BLEU) which is at a small cost of edit distance when compared to CAD and MiCE.

\begin{table}[ht!]
\centering
\small
\begin{tabular}{lcc}
\toprule
\bf Model & \bf self-BLEU $\downarrow$ & \bf Levenshtein $\downarrow$ \\
 \midrule
 \multicolumn{3}{l}{\bf IMDb} \\
CAD & 0.758     & \bf 0.156         \\
MiCE & 0.709 & 0.195 \\
CF-DPR     & \bf 0.002  &   0.830         \\
CORE      & 0.445     & 0.506     \\
\midrule
\multicolumn{3}{l}{\textbf{MNLI Crowd-sourcing  Experiment}} \\
Retrieval & \bf 0.092     & 0.765         \\
w/o Retrieval & 0.313 & \bf 0.484 \\ 
\bottomrule
\end{tabular}
\caption{Closeness (edit distance) and diversity (self-BLEU) for different counterfactual generation strategies}

\label{tab:IMDb_intrinsic}
\end{table}



\begin{table*}[ht!]
\small
\begin{tabular}{llcccccccc}
\toprule
\multicolumn{1}{c}{\bf Dataset} &
\multicolumn{1}{c}{\bf Model} & \multicolumn{1}{c}{\bf Negation} & \multicolumn{1}{c}{\bf Insertion} & \multicolumn{1}{c}{\bf Resemantic} & \multicolumn{1}{c}{\bf Lexical} & \multicolumn{1}{c}{\bf Quantifier} & \multicolumn{1}{c}{\bf Restructure} & \multicolumn{1}{c}{\bf Delete} & \multicolumn{1}{c}{\bf UNK}  \\ \midrule
\multirow{2}{*}{MNLI} & CORE  & 691 & 1303    & 1320   & 1073 & 138 & 92 & 79 & 4577 \\ 
& GPT-3 & 1400 & 705  & 1169   & 2184 & 290  & 95  & 94 &  3545 \\ 
\midrule
\multirow{2}{*}{SST2} & CORE  & 81   & 140  & 202  & 174  & 49 & 5 & 60 &  1305 \\ 
& Polyjuice  & 161 & 91   & 209  & 362 & 26 & 17  & 66 &  1076                                               \\ \bottomrule
\end{tabular}
\caption{List of perturbation types detected using the set of heuristics from \citet{wu-etal-2021-polyjuice}. We can see that for both the task Sentiment and NLI, CORE covers all perturbation types without any explicit control code.}
\label{tab:perturbation_coverage}
\end{table*}

\vspace{-10pt}
\paragraph{Perturbation type}
To further analyze the source of diversity, we classify perturbations in CORE counterfactuals according to the
perturbation-type detector used in \citet{wu-etal-2021-polyjuice}. Table~\ref{tab:perturbation_coverage} shows that CORE is able to cover a broad set of previously defined perturbation types that were recognized in prior work, such as \textit{negation, insertions, lexical change and resemantics} without being explicitly controlled. The number of perturbation types that are not categorized by this detector are significantly higher in case of CORE compared to Polyjuice. In Table~\ref{tab:perturbation_diversity}, we find that CORE perturbations fall in more than one Polyjuice  categories and avoid trivial perturbations like negations and antonyms. 
More examples of the different perturbations we see in CORE are in Appendix~\ref{app:qual_core}

\vspace{-5pt}
\paragraph{Supporting Human Annotation}
\label{results:human_authorin}
Prior work proposes aiding crowd-workers in the task of dataset creation using generative assistants \citep{bartolo-etal-2021-improving} or randomly sampled words \citep{gardner-etal-2021-competency} to encourage creativity, leading to better quality and diversity in human-authored data. We analyze the impact of CF-DPR counterfactuals in encouraging humans to make diverse counterfactual perturbations to text. We design a controlled crowdsourcing  experiment where 200 original MNLI  examples from the validation set are shown to humans with and without the retrieved counterfactual sentences to aid them. 
Human-authored counterfactuals conditioned on retrieved outputs display more diversity, with lower self-BLEU and higher Levenshtein distance compared to the control condition (Table~\ref{tab:IMDb_intrinsic}). 
Qualitative differences between human authored counterfactuals in both conditions and details of the crowd-sourcing are in the Appendix~\ref{appendix:implimentation_details}.





\section{Conclusion}
We present \textbf{CORE}, a retrieval-augmented generation framework for creating \emph{diverse} counterfactual perturbations for CDA. CORE first learns to retrieve relevant text excerpts and then uses GPT-3 few-shot editing conditioned on  retrieved text to make counterfactual edits. 
CORE encourages diversity with the use of additional knowledge for this task, explicitly via retrieval and implicitly (parametric knowledge), via GPT-3 editing.
Conditioning language model edits on naturally occurring data results in diversity
(\S\ref{results:diversity}). 
CORE counterfactuals are more effective at improving generalization to
OOD data compared to other approaches, on natural language inference and sentiment analysis(\S\ref{results:augmentation}). CORE's retrieval framework can also be used to reduce priming effects and encourage diversity in manually authored perturbations (\S\ref{results:instrinsic_eval}).

\section{Limitations}
While CORE involves no human intervention, it is not completely accurate at performing label flipping perturbations and human re-labelling can be beneficial. Our framework uses a learned retriever which can be challenging to train when there are finer semantic differences between classes (e.g. neutral class) that need to be captured. Work on counterfactual generation has focused exclusively on English language text, and it would be an interesting future work to expand such frameworks for other languages.

\section*{Acknowledgements}
We would like to thank Alisa Liu, Rajarshi Das, Sewon Min, Tongshuang Wu, Mandar Joshi, Julian Michael and the anonymous reviewers
for their helpful comments and suggestions.


\bibliography{anthology,custom}
\bibliographystyle{acl_natbib}

\newpage
\appendix

\section{Implimentation details for CORE}
\label{appendix:implimentation_details}

\paragraph{CF-DPR}
We make use of the open source implementation \footnote{\href{https://github.com/facebookresearch/DPR}{https://github.com/facebookresearch/DPR}} of DPR to train the CF-DPR model. We train the CF-DPR model using 2 hard negatives, one the paraphrase and the second the query instance (hypothesis in case of MNLI and entire review for IMDB). All training was performed using 3 TITAN X (Pascal) GPUs. The average run time for CF-DPR for the MNLI task was 5hrs and for IMDB it was 1hr. Total emissions are estimated to be 4.19 kgCO$_2$eq of which 0\% was directly offset. Estimations were conducted using the \href{https://mlco2.github.io/impact#compute}{MachineLearning Impact calculator} presented in \cite{lacoste2019quantifying}.  The training parameters used to train CF-DPR for the MNLI task are in Table ~\ref{tab:cfdpr_mnli} and for IMDB Table~\ref{tab:cfdpr_imdb}. 
\begin{table}[ht!]
    \centering
    \begin{adjustbox}{width=0.4\textwidth}
    \begin{tabular}{c|c}
    \toprule
    \bf Hyperparameters & \bf Value \\
    \midrule
    \midrule
    encoder sequence\_length & 32 \\
    pretrained\_model\_cfg & bert-base-uncased \\
    per\_device\_batch\_size & 32 \\
    weight\_decay & 0.01 \\
    warmup\_steps & 2500 \\
    max_grad\_norm & 2.0 \\
    learning\_rate & 2e-5 \\
    num\_train\_epochs & 10 \\
    eval\_per\_epoch & 10 \\
    hard\_negatives & 2 \\
    \bottomrule
    \end{tabular}
    \end{adjustbox}
    \caption{Hyperparameters for training CF-DPR for counterfactual retrieval on MNLI.}
    \label{tab:cfdpr_mnli}
\end{table}

\begin{table}[ht!]
    \centering
    \begin{adjustbox}{width=0.4\textwidth}
    \begin{tabular}{c|c}
    \toprule
    \bf Hyperparameters & \bf Value \\
    \midrule
    \midrule
    encoder sequence\_length & 96 \\
    pretrained\_model\_cfg & bert-base-uncased \\
    per\_device\_batch\_size & 16 \\
    gradient\_accumulation\_steps & 3 \\
    weight\_decay & 0.01 \\
    warmup\_steps &  100 \\
    max_grad\_norm & 2.0 \\
    learning\_rate & 2e-5 \\
    num\_train\_epochs & 100 \\
    eval\_per\_epoch &  1 \\
    hard\_negatives &  2 \\
    \bottomrule
    \end{tabular}
    \end{adjustbox}
    \caption{Hyperparameters for training CF-DPR for counterfactual retrieval on IMDB.}
    \label{tab:cfdpr_imdb}
\end{table}

We experiment with several hyperparameters and find the quality of the hard negatives used to train the model played the most critical role in training. Adding more hard negatives didn't really help boost performance. Changing learning rate and warmup steps did not drastically affect the training process either. For evaluation we use the validation sets from \citep{kaushik2020learning} and MNLI validation- matched set for sentiment and NLI CF-DPR models respectively and reformat it as described in \S\ref{sec:method_sentiment} and \S\ref{sec:method_nli}. For every instance in the validation set, we generate 30 random negatives and 30 hard negatives. Since we cannot generate 30 different paraphrases we make use of a semantic sentence retriever model \citep{gao-etal-2021-simcse} to retrieve semantically similar sentences from a text corpus. Finally following the DPR codebase we measure the top 1 accuracy on the validation set.  For MNLI our CF-DPR model achieves an top 1 accuracy of about 70\% while for IMDb around 45\%.
\par The cross encoder model was trained using the \texttt{sentence-transformers}\footnote{\href{https://github.com/UKPLab/sentence-transformers}{https://github.com/UKPLab/sentence-transformers}} library. Each instance in the training dataset for MNLI contains a query instance which is a concatenated version of the premise and hypothesis separated by the [SEP] token, and two hypothesis with labels 1 and 0 respectively. We train the \texttt{bert-base-model} using the BCE loss. We train the model for 10 epochs with a batch size of 64 and learning rate $2e-5$. We evaluate every 2000 setps and save the model with the best evaluation loss. The cross encoder gave a validation accuracy of 91\%.

For Inference we encode the entire search corpus with the context encoder and index it using faiss. We approximate maximum inner-product search (MIPS) with an Inverted File Index (IVF) \citep{1238663} for faster retrieval. We use the IVFFlat index \footnote{\href{https://github.com/facebookresearch/faiss/wiki/Faiss-indexes}{https://github.com/facebookresearch/faiss/wiki/Faiss-indexes}} as it helps improve the retrieval speed at a small cost of accuracy. We set the number of centroids (K) as 300 and n\_probes to 30.


\paragraph{GPT-3 few shot prompting}
We use the \texttt{text-davinci-002} model as the Editor. The prompts to GPT-3 for IMDB is in Table ~\ref{tab:gpt3_prompt_imdb} and for MNLI Table~\ref{tab:gpt3_prompt_mnli}. For both the model settings we set the temperature parameter to $0.7$  and Top p parameter to $1$.
\begin{table*}[ht!]
    \centering
    \small
    \adjustbox{max width=\textwidth}{
    \begin{tabular}{p{1\linewidth}}
    \toprule
         Two sentences are given, sentence 1 and sentence 2. Given that Sentence 1 is True, Sentence 2 (by implication), must either be (a) definitely True, (b) definitely False. You are presented with an initial Sentence 1 and Sentence 2 and the correct initial relationship label (definitely True or definitely False). Edit Sentence 2 making a small number of changes such that the following 3 conditions are always satisfied: \\
         (a) The target label must accurately describe the truthfulness of the modified Sentence 2 given the original Sentence 1. \\
         (b) In order to edit sentence 2, must only make use of one or two relevant words present in the provided list of words. \\
         (c) Do not rewrite sentence 2 completely, only a small number of changes need to be made. Here are some examples:\\
         original sentence 1: Oh, you do want a lot of that stuff.  original sentence 2: I see, you want to ignore all of that stuff.    initial relationship label: definitely False \\
         target label: definitely True \\
         List of words: ['correct', 'before', 'happen', 'already', 'more', 'saw', 'on'] \\
         modified sentence 2: I see, you want more of that stuff.
         original sentence 1: Region wide efforts are also underway.    original sentence 2: Regional efforts have not stopped. initial relationship label: definitely True \\
         target label: definitely False \\
         List of words: ['talking', 'hold', 'put', 'caller', 'on'] \\
         modified sentence 2: Regional efforts are on hold at the moment. \\
         original sentence 1: yeah you could stand in there if you really wanted to i guess. original sentence 2: If you want you can sit there I guess. initial relationship label: definitely False \\
         target label: definitely True \\
         List of words: ['there', 'without', 'ants', 'stand', 'even'] \\
         modified sentence 2: If you want you can stand there I guess. \\
         original sentence 1: and uh every every opportunity there is to make a dollar he seems to be exploiting that.    original sentence 2: He works to make money. initial relationship label: definitely True \\
         target label: definitely False \\
         List of words: ['work', 'pass', 'dollars', 'up', 'owe'] \\
         modified sentence 2: He passes up on opportunities to make money. \\
         \bottomrule
    \end{tabular}}
    \caption{Prompt given to GPT-3 for generating CORE counterfactuals for MNLI. The instructions are similar to the ones given to crowdworkers in \citep{kaushik2020learning}, additionally we incorporate 4 demonstrations. The test instance is appended to this prompt.}
    \label{tab:gpt3_prompt_mnli}
\end{table*}

\begin{table*}[ht!]
    \centering
    \small
    \begin{adjustbox}{width=\textwidth}
    \begin{tabular}{p{1\linewidth}}
    \toprule
    Given a movie review and its sentiment. Edit the review making a small number of changes such that, the following two conditions are always satisfied: \\
    (a) the target label accurately describes the sentiment of the edited review\\
    (b) make use of only a few key words provided in the list of words to edit the review\\Do not remove or add any extra information, only make changes to change the sentiment of the review.\\ Review: Long, boring, blasphemous. Never have I been so glad to see ending credits roll.\\ Label: Negative\\ List of relevant words: ['clean' , 'now', 'brought', 'perfect' , 'many' ,'interesting' ,'memories', 'sad']\\ Target Label: Positive\\ Edited Review: Perfect, clean,interesting. Never have I been so sad to see ending credits roll.\\ Review: I don't know why I like this movie so well, but I never get tired of watching it.\\ Label: Positive\\ List of relevant words: ['hate' , 'now', 'supposedly' , 'many' , 'memories', 'watching']\\ Target Label: Negative\\ Edited Review: I don't know why I hate this movie so much, now I am tired of watching it. \\
    \bottomrule
    \end{tabular}
    \end{adjustbox}
    \caption{Prompt given to GPT-3 for generating CORE counterfactuals for IMDB. The instructions are similar to the ones given to crowdworkers in \citep{kaushik2020learning}, additionally we incorporate 2 demonstrations. The test instance is appended to this prompt.}
    \label{tab:gpt3_prompt_imdb}
\end{table*}

\paragraph{DeBERTa finetuning}
All training was performed using 3 TITAN X (Pascal) GPUs. We evaluate every 400 steps and save the model with the best evaluation loss. We make use of the huggingface trainer APIs ~\footnote{\href{https://github.com/huggingface/transformers}{https://github.com/huggingface/transformers}} for fine-tunning the models. It takes around 2 hrs to fine-tune the DeBERTa base model for MNLI and less than an hour for IMDb. The fine-tuned models were evaluated on publicly available validation/test sets. 
\par For MNLI in-domain performance we only report results on the mis-matched validation set as we observe that both the sets matched and mis-matched had similar scores. For QNLI, SNLI we use the datasets that are part of the GLUE benchmark. For WaNLI, HANS and Diagonistics we use the same set of validation sets used in \citet{liu2022wanli} and for ANLI and LI\_LI we use the official datasets provided by the authors. Since HANS and QNLI are binary NLI tasks (entailment and non-entailment), for measuring accuracy we consider both Neutral and Contradiction predictions as non-entailment.
\par For IMDb we use all the official datasets available. For CAD we combine the validation and tests in-order to get more statistically significant results.
\begin{table}[ht!]
    \centering
    \begin{tabular}{c|c}
    \toprule
    \bf Hyperparameters & \bf Value \\
    \midrule
    \midrule
    Model & microsoft/deberta-base \\
    learning rate & $2e^{-5}$\\
    number of epochs & 3 \\
    per device batch size & 32 \\ 
    max seq length  & 128 \\
    \bottomrule
    \end{tabular}
    \caption{Training Hyperparameters for DeBERTa base for MNLI.}
    \label{tab: training_details_mnli}
\end{table}

\begin{figure*}
    \centering
    \includegraphics[width=\textwidth]{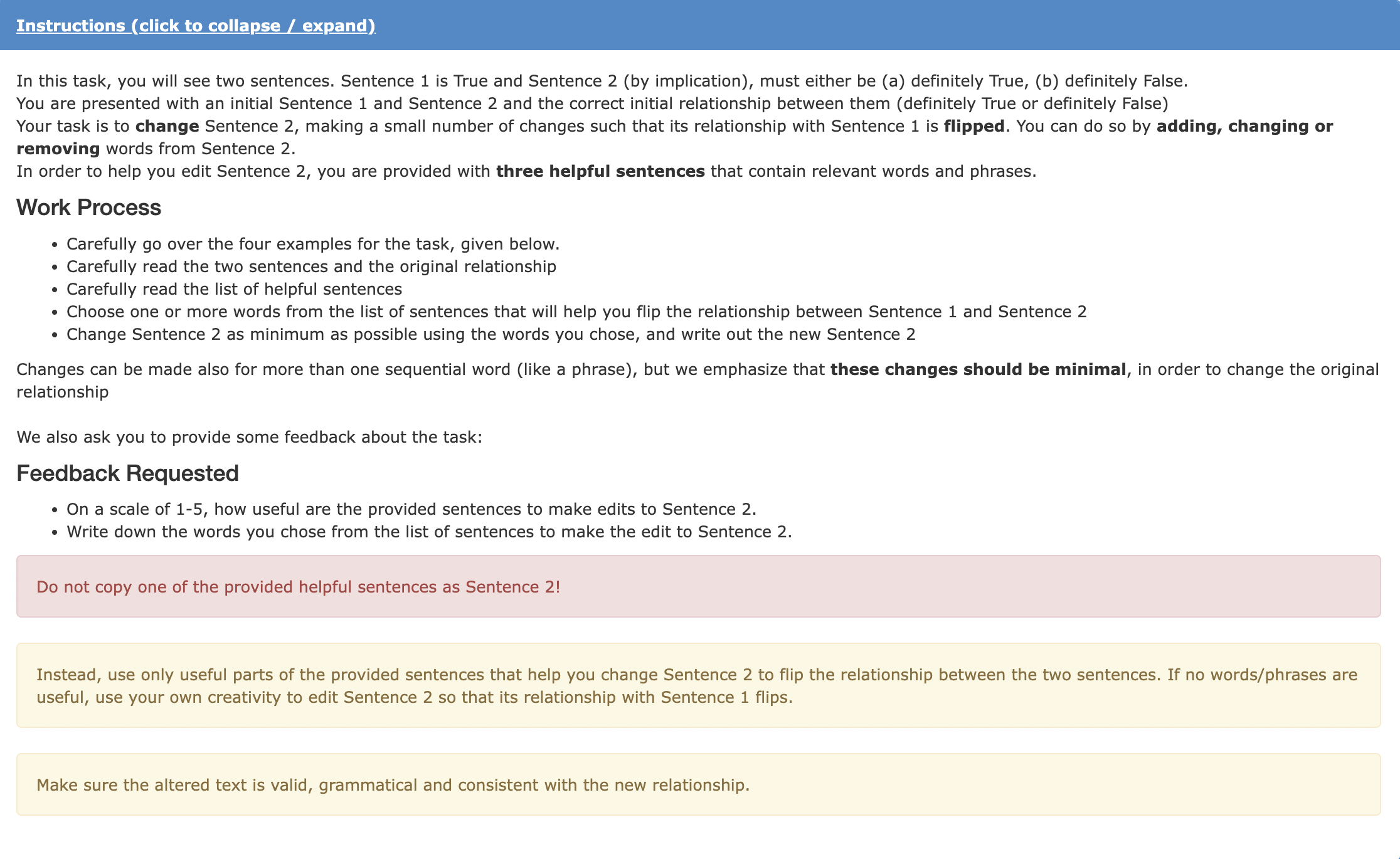}
    \includegraphics[width=\textwidth]{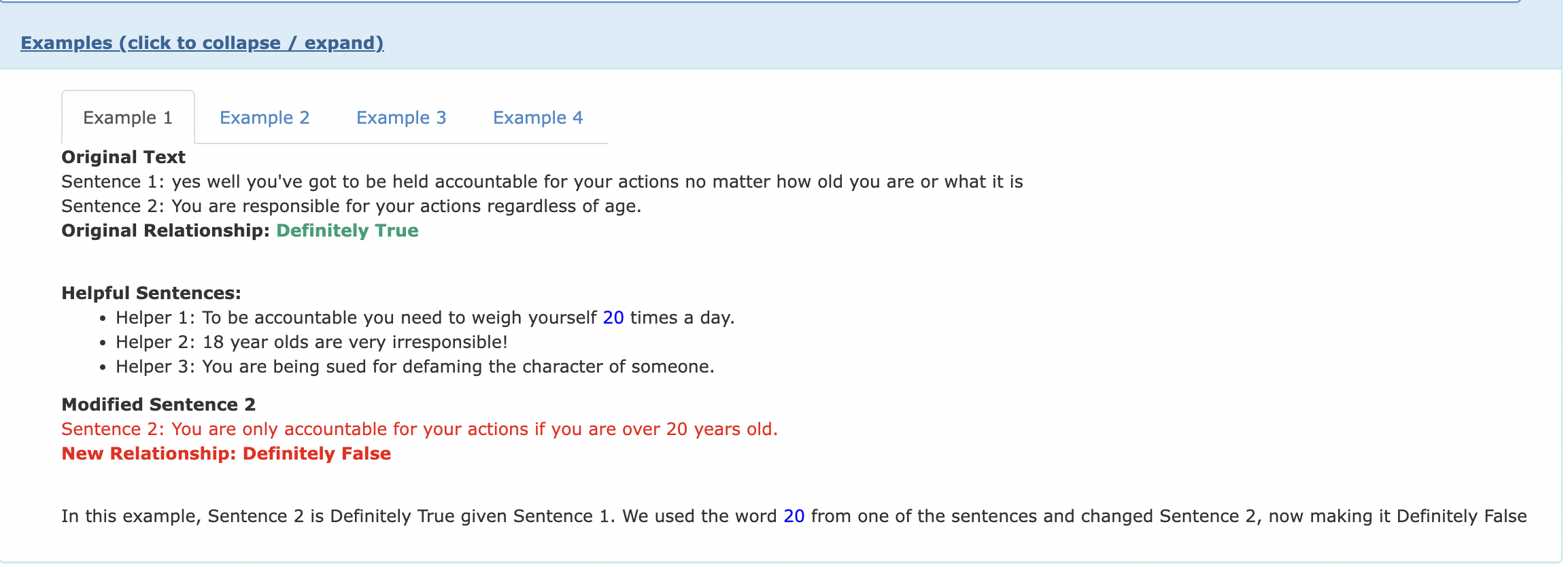}
    \caption{Instructions to crowd-workers for the counterfactual editing task conditioned on retrieval.}
    \label{fig:instructions}
\end{figure*}

\begin{table}[ht!]
    \centering
    \begin{tabular}{c|c}
    \toprule
    \bf Hyperparameters & \bf Value \\
    \midrule
    \midrule
    Model & microsoft/deberta-base \\
    learning rate & $2e^{-5}$\\
    number of epochs & 5 \\
    per device batch size & 32 \\ 
    max seq length  & 128 \\
    \bottomrule
    \end{tabular}
    \caption{Training Hyperparameters for DeBERTa base for IMDB.}
    \label{tab: training_details_imdb}
\end{table}
\paragraph{Crowd-sourcing Study}
In Section \ref{results:human_authorin}, we use CF-DPR outputs to aid crowd-sourcing of counterfactual edits. On the MNLI development set, we randomly sample 200 instances. We create two user interfaces for crowdworkers on Amazon Mechanical Turk to collect data under two conditions. In one condition, workers are shown the original instance (premise and hypothesis) and top-three retrieved counterfactuals provided by CF-DPR. They are asked to edit the hypothesis following brief instructions instructions and examples, shown in Figure~\ref{fig:instructions}. In the second case, they are just shown the original instance, no retrieved outputs.  When revising
examples, we asked workers to preserve the intended meaning  through minimal revisions. Each instance is modified only once and different annotators are shown instances from both sets. 
Annotators were required to have a HIT approval
rate of 90\%, a total of 1,000 approved HITs.
For the case where annotators were shown retrieved sentences, we found that annotator quality was quite poor, since annotators were not filtered by  a qualification test to do the task. More generally, complex annotator tasks often require substantial training of crowd-workers \citep{bartolo-etal-2021-improving}, which is outside the scope of this work. Instead, we recruit computer science graduate students (outside of the study) to get annotations for this task.

\paragraph{Compensation} We aimed to pay rate of
at least $\$15$ per hour. Workers were paid $0.75$ for
each example that they annotate.

\section{Qualitative Examples - Human Authored Counterfactuals}
\label{app:qual_core}
We show examples of the counterfactual editing of MNLI examples done by crowd-workers who were asked to independently edit instances and experts (CS graduate students) who were shown CF-DPR-retrieved counterfactual instances in Table~\ref{tab:crowdworker_examples}.

\begin{table}[h!]
    \centering
    \small
    \begin{tabular}{llll}
         \toprule
         \multicolumn{2}{c}{\bf CAD} & \multicolumn{2}{c}{\bf CORE} \\
         \bf Biased feature &  \bf z\_stats & \bf Biased feature &  \bf z\_stats \\
         \midrule
         \multicolumn{3}{l}{\textbf{Negative}} \\
         
         bad & 16.93 & unfortunately & 12.27 \\
         worst & 16.71 & worst & 10.83 \\
         terrible & 15.44 & terrible & 10.55 \\
         boring & 15.05 & bad & 10.49 \\
        \midrule
        \multicolumn{3}{l}{\textbf{Positive}} \\
        great & 19.41 & great & 11.15 \\
        best & 11.54 & best & 8.04 \\
        amazing & 11.25 & surprisingly & 7.66 \\
        wonderful &9.47 & wonderful & 4.73 \\
        \bottomrule

    \end{tabular}
    \caption{Z scores values for the original $1.7k$ imdb reviews augmented either with $1.7k$ CAD or CORE counterfactuals.}
    \label{tab:app_z_stats_imdb}
\end{table}

In addition to the examples present in Table~\ref{tab:example_table} we also provide some more examples for MNLI in Table~\ref{tab:app_example_table1} and IMDb in Table~\ref{tab:app_example_table2}.

\section{Additional Analysis}
\label{appendix:additional_results}

\paragraph{Z statistics scores}
Figure~\ref{fig:z_stats} depicts the competancy style plot for the IMDb augmented subset. In addition to that we also show the individual z scores in Table~\ref{tab:app_z_stats_imdb}
\paragraph{SST-2 Results}
We also study the ability of CORE to generate counterfactuals on SST2 without being explicitly trained on SST2 data. We use the CF-DPR model trained on IMDb. The results and comparison with Polyjuice can be found in Table~\ref{sst2_results}.

\paragraph{CDA baselines}
For all the baselines we make use of the official implementations. For Tailor on MNLI, we use both no context and in-context \texttt{swap-core} perturbations.

\begin{table*}[ht!]
    \centering
    \resizebox{1\textwidth}{!}{
    \begin{tabular}{l | g | k | k | k | b | b |  b|  b b  b}
    \toprule
      \multirow{2}{*}{\bf Train $\downarrow$ \bf Test $\rightarrow$} & \cellcolor[gray]{0.9}  & 
      & 
      & 
      & 
       & 
    &  
      \bf NLI-Adv & \multicolumn{3}{| b}{\bf ANLI} \\
      
     & \multirow{-2}{*}{\cellcolor[gray]{.9}\bf MNLI} 
     & \multirow{-2}{*}{\bf QNLI}
     & \multirow{-2}{*}{\bf SNLI} 
     & \multirow{-2}{*}{\bf WaNLI}
     & \multirow{-2}{*}{\bf HANS}
     & \multirow{-2}{*}{\bf Diagno}
     & \bf \small LI_LI & \bf R1 & \bf R2 & \bf R3 \\
    \midrule
    MNLI 
    & 87.66 
    & 50.57
    & 83.82
    & 59.10
    & 68.22
    & 61.11
    & 90.39
    & 32
    & 30
    & 28.58\\
~~~~~ + TF-IDF + GPT3 Editor 
& 87.89
& 50.57
& 83.13
& 58.86
& 60.51
&  61.77
& 87.49
& 32.30
& 29.50
& 30.41
\\
~~~~~ + CORE counterfactuals 
& 87.97
& 50.52
& 84.34
& \bf 60.80
& \bf 72.57
& 62.32
& 90.98
& 33 
& 28.90
& \bf 30.58
\\

    \bottomrule

    \end{tabular}}
    \caption{Accuracies of data augmentation for CORE and retrival ablation. We observe that using a simpler TF-IDF based retriver doesn’t help improve performance across any task, in contrast to using a counterfactual retriever.}
    
    \label{ablation_mnli_results}
\end{table*}
\begin{table*}[ht!]
\small
\begin{tabular}{lllllll}
\toprule
\bf Train $\downarrow$ \bf Test $\rightarrow$   & \bf SST2         & \bf Senti140    & \bf Yelp               & \bf IMDb         & \bf IMDb contrast        & \bf IMDb cad            \\
\midrule
SST2   (6k)         & 89.60 (0.46) & 75.4 (1)    & 86.75 (0.6)        & 80.43(0.88)  & 78.27 ( 1.37)        & 83.2 (1.3)          \\
SST2 (4k)+ Polyjuice (2k) & 89.64 (0.67) & 75.9 (0.49) & 85.5 (0.12)        & 81.10 (0.35) & \textbf{84.7 (0.36)} & \textbf{88.2 (0.4)} \\
SST2 (4k) + CORE (2k)      & 88.45 (1)    & 75.1 (0.18) & \textbf{87 (0.87)} & 81.75 (0.8)  & 80.19 (0.09)         & 84.72 (0.38)    \\   
\bottomrule
\end{tabular}
\caption{Performance of CORE on SST2. We can see that although our CORE framework has not been trained to generate counterfactuals for SST2 it can yet achieve compatible scores as Polyjuice on out-of-distribution test sets. }
\label{sst2_results}
\end{table*}

\begin{table*}[ht!]
    \centering
\begin{tabular}{p{0.45\linewidth}|p{0.45\linewidth}}
\toprule
\bf without Retrieval  & \bf with Retrieval\\
\midrule 
\multicolumn{2}{p{0.90\linewidth}}{\textbf{Premise}: Another book that I read recently is very interesting books The Journals of Lewis and Clark. \textbf{Hypothesis}: I recently read The Journals of Lewis and Clark. (Entailment)} \\ 
\textbf{New Hypothesis}: I skipped reading The Journals of Lewis and Clark. & \textbf{New Hypothesis}: I've never heard of this book The Journals of Lewis and Clark.\\
\midrule
\multicolumn{2}{p{0.90\linewidth}}{\textbf{Premise}: This northern beach of magnificent tan sand is most agreeably reached by boat. \textbf{Hypothesis}:The beach has beautiful sand. (Entailment)} \\
\textbf{New Hypothesis}: The sand of the beach is ordinary & \textbf{New Hypothesis}: The beach is impossible to reach by boat and the volcano has devastated the island. \\
\midrule
\multicolumn{2}{l}{\textbf{Premise}: You never call. \textbf{Hypothesis}: You rarely call on the phone, nor webcam. (Entailment)} \\
\textbf{New Hypothesis}: You always call or webcam & \textbf{New Hypothesis}: You run up a big bill on your phone calling me! \\
\midrule 
\multicolumn{2}{p{0.90\linewidth}}{\textbf{Premise}:have that well and it doesn't seem like very many people uh are really i mean there's a lot of people that are on death row but there's not very many people that actually um do get killed \textbf{Hypothesis}:  There are a lot of people on death row, but not that many actually are executed. (Contradiction) }\\
\textbf{New Hypothesis}: There are not many people on death row, because most are promptly executed. & \textbf{New Hypothesis}: There are a lot of people on death row, and thousands get killed without anyone noticing. \\
\midrule
\multicolumn{2}{p{0.90\linewidth}}{\textbf{Premise}: Generally, if pH of scrubbing liquor falls below a range of 5.0 to 6.0, additional reagent is required to maintain the reactivity of the absorbent. \textbf{Hypothesis}: if pH of scrubbing liquor falls below a range of 5.0 to 6.0, then the whole world may explode (Contradiction)} \\
\textbf{New Hypothesis}: If the pH of scrubbing liquor falls below a range of 5.0 to 6.0 more reagent is needed to maintain it's reactivity & \textbf{New Hypothesis}: if pH of scrubbing liquor falls below a range of 5.0 to 6.0, additional reagent will be needed to make the absorbent last longer\\
\midrule
\multicolumn{2}{p{0.90\linewidth}}{\textbf{Premise}: If any of us at the dental school can be of assistance, please write or call. \textbf{Hypothesis}: The dental school may or may not be able to help. (Contradiction)} \\
 \textbf{New Hypothesis}: The dental school cannot give any help. & \textbf{New Hypothesis}: We at the dental school at ready to help, please give us a call.\\
\bottomrule
\end{tabular}
    \caption{Qualitative differences between counterfactual edits made by humans when they are shown retrieved data vs. when they are not. Human annotators primed with retrieval are less likely to use trivial heuristics like negation, antonyms etc, leading to less bias introduced in the new data.}
    \label{tab:crowdworker_examples}
\end{table*}
\begin{table*}[ht!]
    \begin{adjustbox}{width=0.97\textwidth}
    \begin{tabularx}{\linewidth}{p{0.30\linewidth}|p{0.30\linewidth}|p{0.35\linewidth}}
    \toprule
    \multicolumn{3}{c}{\textbf{~~~ MNLI}} \\
    \midrule
    \multicolumn{3}{l}{\textbf{Premise}: this is my first call because i just got my password,} \\
    \multicolumn{3}{l}{\textbf{Hypothesis}: I just got my password, I've never called before.} \\
    \multicolumn{3}{l}{\textbf{Label}: Entailment} \\
    \midrule
    
    \textbf{CFDPR} & \textbf{GPT3} & \textbf{CORE} \\
    I have initiated multiple calls myself &  I just got my password, I've called many times before. &  I've initiated multiple calls before. \\
    
    \midrule
    \multicolumn{3}{l}{\textbf{Premise}: This is routine, he said. }\\
    \multicolumn{3}{l}{\textbf{Hypothesis}: This hardly ever happens, he said} \\
    \multicolumn{3}{l}{\textbf{Label}: Contradiction} \\
    \midrule
    
    \textbf{CFDPR} & \textbf{GPT3} & \textbf{CORE} \\
      He brought a practice routine  & This is not routine, he said.  &  This is brought up in practice, he said\\
      
    \midrule
    \multicolumn{3}{l}{\textbf{Premise}: Neither exercise is intended to revive the patient.} \\
    \multicolumn{3}{l}{\textbf{Hypothesis}:The patient has no exercises to do.} \\
    \multicolumn{3}{l}{\textbf{Label}: Contradiction} \\
    \midrule

    \textbf{CFDPR} & \textbf{GPT3} & \textbf{CORE} \\
    The exercises may revive the patient.
 &  The patient has no exercises to revive them. & The patient has some exercises to do. \\

    \midrule
    \multicolumn{3}{l}{\textbf{Premise}: Many border collie breeders, for example, take great exception to the dog industry's emphasis} \\ 
    \multicolumn{3}{l}{on ideal appearance rather than behavior.} \\
    \multicolumn{3}{l}{ \textbf{Hypothesis}: Border collie breeders don't like that the dog industry cares so much on personality.} \\
    \multicolumn{3}{l}{\textbf{Label}: Entailment} \\
    \midrule

    \textbf{CFDPR} & \textbf{GPT3} & \textbf{CORE} \\
     Border collies have been the most popular dogs this decade cause of its looks. & 
    Border collie breeders don't like that the dog industry doesn't care so much on looks. &
    Most border collie breeders don't like that the dog industry cares so much on looks.\\

    \midrule
    \multicolumn{3}{l}{\textbf{Premise}: Upon Tun Abdul Razak's death in 1976, the post of prime minister was taken up by} \\
        \multicolumn{3}{l}{Datuk Hussein Onn, a son of the founder of the UMNO.} \\
    \multicolumn{3}{l}{ \textbf{Hypothesis }:Datuk Hussein Onn died in 1976, and Tun Abdul Razak became prime minister. } \\
    \multicolumn{3}{l}{\textbf{Label}: Contradiction} \\
    \midrule

    \textbf{CFDPR} & \textbf{GPT3} & \textbf{CORE} \\
    Singapore offered the first prime minister a spot in its federation. 
    & Datuk Hussein Onn succeeded Tun Abdul Razak as prime minister in 1976.
     & Upon Tun Abdul Razak's death in 1976, the first spot of prime minister was offered to Datuk Hussein Onn.\\
     
    \midrule
    \multicolumn{3}{l}{\textbf{Premise}: But he had seen enough.} \\ 
    \multicolumn{3}{l}{ \textbf{Hypothesis}: He didn't need to see any more.} \\
    \multicolumn{3}{l}{\textbf{Label}: Entailment} \\

    \midrule
    
    \textbf{CFDPR} & \textbf{GPT3} & \textbf{CORE} \\
    He still needed to see more.
    & He needed to see more.
    & He still needed to see more.\\
    
     \bottomrule
    \end{tabularx}
    \end{adjustbox}
    \caption{CORE generated counterfactuals for MNLI, along with the CF-DPR retrieved outputs and the independent GPT-3 Editor. We can see that CORE uses diverse words to generate the counterfactuals in compared to the vanilla GPT-3 model. }
    \label{tab:app_example_table1}
\end{table*}

\begin{table*}[ht!]
    \centering
    \small
    \begin{adjustbox}{width=\textwidth}
    \begin{tabularx}{\linewidth}{p{0.23\linewidth}|p{0.36\linewidth}|p{0.34\linewidth}}
    \toprule
    \multicolumn{3}{c}{\textbf{IMDb}} \\
    \midrule
    \multicolumn{3}{p{0.97\textwidth}}{\textbf{Review:} This is one of the worst movies I saw! I dunno what are the reasons for shoting suck a crap. Don't waste your time watching this. Good actors, but extremely bad screenplay and dialogues. Hope there'll be no Blanche 2 :-) Avoid this movie by all means}\\
            \multicolumn{3}{l}{\textbf{Sentiment}: Negative} \\
    \midrule
    \bf CF-DPR & \bf GPT3 & \bf CORE  \\
    One of the best things I have seen in my life. The screne tests are not only a nice piece of film history, but a testament to the positive evolution of the screenplay. That will set the stage.
    & This is one of the best movies I saw! I dunno what are the reasons for shoting such an amazing film. Don't waste your time watching anything else. Good actors, but extremely good screenplay and dialogues. Hope there'll be a Blanche 2 :-) Avoid any other movie by all means!
    & This is one of the best movies I've seen! I dunno what are the reasons for shoting such a great film. Don't waste your time avoiding this. Good actors, but extremely good screenplay and dialogues. Hope there'll be a Blanche 2 :-) Watch this movie by all means!", "old_label": "Negative

    \end{tabularx}
    \end{adjustbox}
    \small
    \begin{adjustbox}{width=1\textwidth}
    \begin{tabularx}{\linewidth}{p{0.23\linewidth}|p{0.36\linewidth}|p{0.34\linewidth}}
    \midrule
    \multicolumn{3}{p{0.97\textwidth}}{\textbf{Review:}Long, boring, blasphemous. Never have I been so glad to see ending credits roll.}\\
    \multicolumn{3}{l}{\textbf{Sentiment}: Negative} \\
    \midrule
    \bf CF-DPR & \bf GPT3 & \bf CORE  \\
     A clean one for now a days. This brought back so many memories!
     & Short, exciting, and full of faith. I was so glad to see the ending credits roll! & 
    Back in the days, this movie was brought to perfection. Clean memories and many interesting ones. 
    \\
    \end{tabularx}
    \end{adjustbox}
     \small
    \begin{adjustbox}{width=\textwidth}
    \begin{tabularx}{\linewidth}{p{0.23\linewidth}|p{0.36\linewidth}|p{0.34\linewidth}}
    \midrule
    \multicolumn{3}{p{0.97\textwidth}}{\textbf{Review:} \"A Bug's Life\" is like a favorite candy bar -- it's chock-full of great little bits that add up to something really tasty. The story couldn't have been better; it's clever, has \"heart\" (emotion), and every character has a nice \"arc\" (a growth or change). By comparison, the only characters in \"Toy Story\" to have an \"arc\" are Buzz, who learns to love being a toy, and Woody, who overcomes his resentment of Buzz. There are tons of laughs and cute moments in \"A Bug's Life\". All of the actors turn in great voice work, and the animation, both the motion and detail, is superb. This serious movie buff doesn't throw around \"10\"s lightly, but this movie certainly deserves the \"10\" I gave it.}\\
    \multicolumn{3}{l}{\textbf{Sentiment}: Positive } \\
    \midrule
    \bf CF-DPR & \bf GPT3 & \bf CORE  \\
    That is rare. Lives up to it's name. This, originally, wasn't one of them. This, unfortunately, is not one of them. But no laughs. 
    However, it really only warrants one. & \"A Bug's Life\" is like a favorite candy bar -- it's chock-full of great little bits that add up to something really distasteful. The story couldn't have been worse; it's dull, has \"heart\" (emotion), and every character has a nice \"arc\" (a growth or change). By comparison, the only characters in \"Toy Story\" to have an \"arc\" are Buzz, who learns to love being a toy, and Woody, who overcomes his resentment of Buzz. There are tons of laughs and cute moments in \"A Bug's Life\". All of the actors turn in great voice work, and the animation, both the motion and detail, is superb. This serious movie buff doesn't throw around \"10\"s lightly, but this movie certainly deserves the \"1\" I gave it.
    &
    A Bug's Life is like a favorite candy bar -- it's chock-full of great little bits that add up to something really tasty. Unfortunately, the story couldn't have been worse; it's d- grade, has no "heart" (emotion), and every character is a one-dimensional flatline. By comparison, the only characters in \"Toy Story\" to have an \"arc\" are Buzz, who learns to love being a toy, and Woody, who overcomes his resentment of Buzz. There are none of laughs and cute moments in A Bug's Life. All of the actors turn in terrible voice work, and the animation, both the motion and detail, is terrible. This serious movie buff doesn't throw around "10"s lightly, but this movie certainly doesn't deserve the 10 I gave it.\\
    \bottomrule
    \end{tabularx}
    \end{adjustbox}
    \caption{CORE generated counterfactuals for IMDb, along with the CF-DPR retrieved outputs and the independent GPT-3 Editor. The CFDPR results are the concatenated set of retrived sentences.}
    \label{tab:app_example_table2}
\end{table*}

\begin{figure*}
    \centering
    \includegraphics[width=\textwidth]{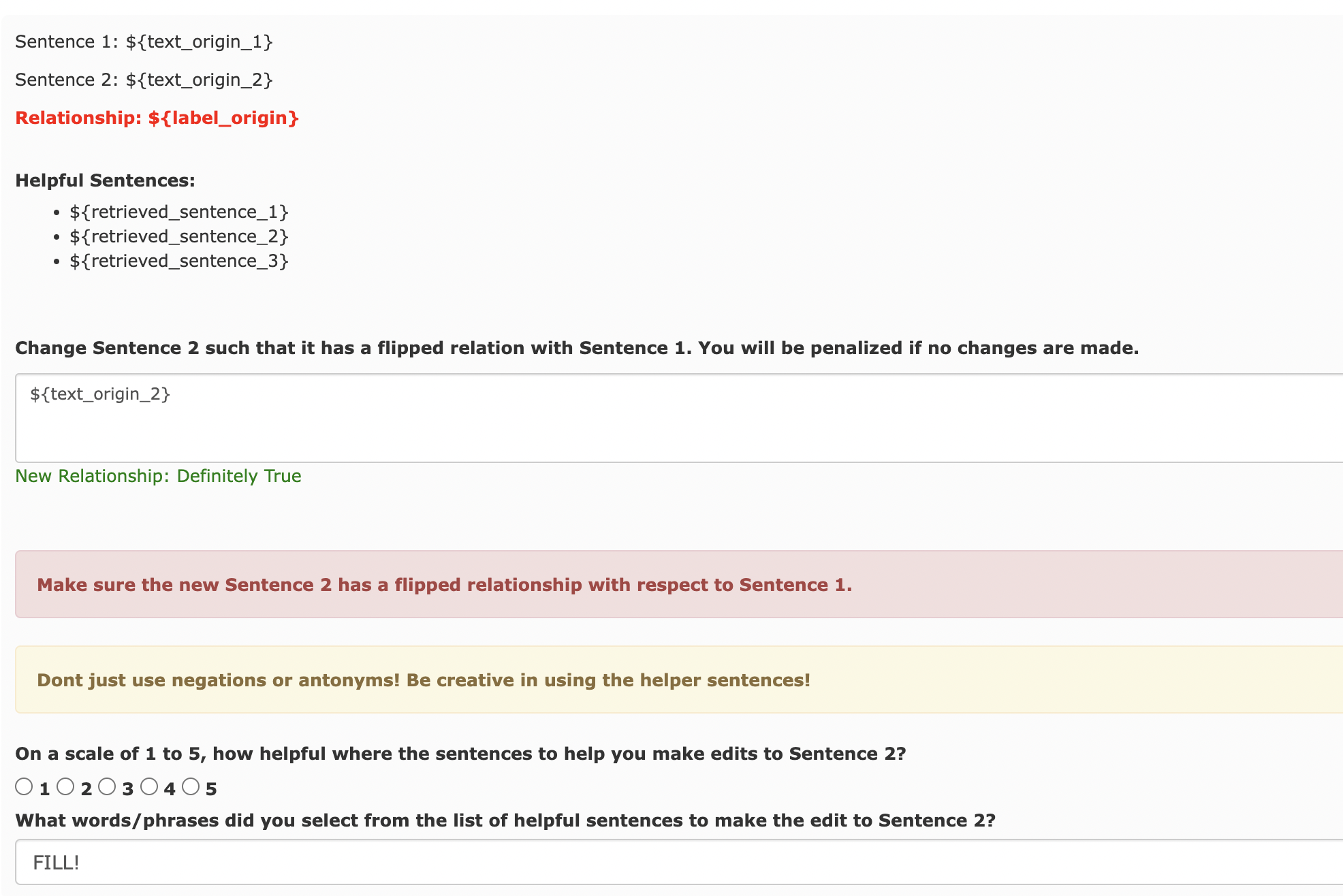}
    \caption{Crowd-worker platform interface where humans have to use retrieved sentences to edit examples}
    \label{fig:with_retrieval}
\end{figure*}

\begin{figure*}
    \centering
    \includegraphics[width=\textwidth]{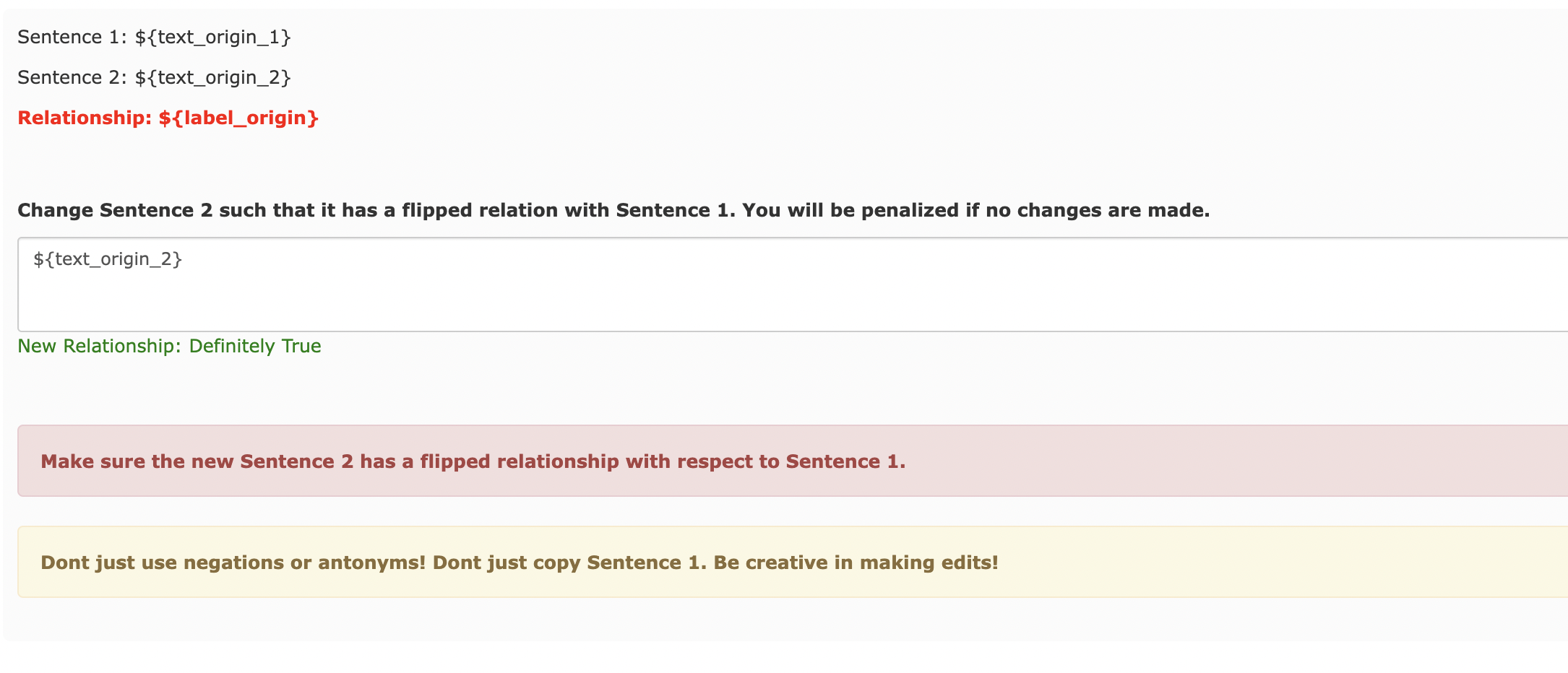}
    \caption{Crowd-worker platform interface where humans  have to edit examples without any priming}
    \label{fig:without_retrieval}
\end{figure*}

\label{sec:appendix}


\end{document}